\documentclass[11pt]{article}

\usepackage[top=1in, bottom=1in, left=1in, right=1in]{geometry}
\usepackage{lineno}
\usepackage[numbers,sort&compress]{natbib}
\usepackage{amssymb}
\usepackage{amsmath}
\usepackage{amsthm}
\usepackage{epsfig}
\usepackage{graphicx}
\usepackage{graphics}
\usepackage{float}
\usepackage{subfigure}
\usepackage{multirow}
\usepackage{color}
\usepackage{lineno}
\usepackage{fullpage}
\usepackage[normalem]{ulem} 
\usepackage{makeidx}
\usepackage{xspace}
\usepackage{wrapfig}
\usepackage{hyperref}
\usepackage{algorithm, algcompatible}
\usepackage{algpseudocode}
\usepackage{setspace}
\usepackage[nameinlink,capitalise]{cleveref}

\let\oldReturn\Return
\renewcommand{\Return}{\State\oldReturn}

\definecolor{darkred}{rgb}{1, 0.1, 0.3}
\definecolor{darkblue}{rgb}{0.1, 0.1, 1}
\definecolor{darkgreen}{rgb}{0,0.6,0.5}

\newcommand {\mm}[1] {\ifmmode{#1}\else{\mbox{\(#1\)}}\fi}

\algnewcommand\algorithmicinput{\textbf{Input:}}
\algnewcommand\INPUT{\item[\algorithmicinput]}
\algnewcommand\algorithmicoutput{\textbf{Output:}}
\algnewcommand\OUTPUT{\item[\algorithmicoutput]}
\algnewcommand\algorithmicoptional{\textbf{Optional:}}
\algnewcommand\OPTIONAL{\item[\algorithmicoptional]}

\providecommand{\keywords}[1]
{
  \small	
  \textbf{\textit{Keywords:}} #1
}
\begin{document}
\title{\vspace{-.3in}Hierarchical Deep Learning of Multiscale Differential Equation Time-Steppers\vspace{-.05in}}
\author{
{Yuying Liu$^{\dag}$\footnote{email: yliu814@uw.edu}, J. Nathan Kutz$^{\dag,\ddag}$, Steven L. Brunton$^{\dag,\ddag}$}\\[.1in]
{$^{\dag}$ {\small Department of Applied Mathematics, University of Washington, Seattle, WA 98105}}\\
{$^{\ddag}${\small  Department of Mechanical Engineering, University of Washington, Seattle, WA 98105\vspace{-.35in}}}
}
\date{}
\maketitle
\setcounter{page}{1}

\begin{abstract}
Nonlinear differential equations rarely admit closed-form solutions, thus requiring numerical time-stepping algorithms to approximate solutions. 
Further, many systems characterized by multiscale physics exhibit dynamics over a vast range of timescales, making numerical integration computationally expensive due to numerical stiffness.  
In this work, we develop a hierarchy of deep neural network time-steppers to approximate the flow map of the dynamical system over a disparate range of time-scales.  
The resulting model is purely data-driven and leverages features of the multiscale dynamics,  enabling numerical integration and forecasting that is both accurate and highly efficient.  
Moreover, similar ideas can be used to couple neural network-based models with classical numerical time-steppers.  Our {multiscale} hierarchical time-stepping scheme provides important advantages over current time-stepping algorithms, including (i)  circumventing numerical stiffness due to disparate time-scales, (ii) improved accuracy in comparison with leading neural-network architectures, 
(iii)  efficiency in long-time simulation/forecasting due to explicit training of slow time-scale dynamics, 
and (iv) a flexible framework that is parallelizable and may be integrated with standard numerical time-stepping algorithms.   
The method is demonstrated on a wide range of nonlinear dynamical systems, including the Van der Pol oscillator, the Lorenz system, the Kuramoto–Sivashinsky equation, and fluid flow pass a cylinder; audio and video signals are also explored. 
On the sequence generation examples, we benchmark our algorithm against state-of-the-art methods, such as LSTM, reservoir computing, and clockwork RNN.  
Despite the structural simplicity of our method, it outperforms competing methods on numerical integration. 
\end{abstract}

\section{Introduction}
\label{sec:intro}

Scientific computing has revolutionized nearly every scientific discipline, allowing for the ability to model, simulate, engineer, and optimize a complex system's design and performance.  
This capability has been especially important in nonlinear, multiscale systems where recourse to analytic and perturbation methods are limited.  
For instance, modern high-fidelity simulations enable researchers to design aircraft, simulate the evolution of galaxies, quantify atmospheric and ocean interactions for weather forecasting, and model high-dimensional neuronal networks of the brain.  
Thus, given a set of governing equations, typically spatio-temporal {\em partial differential equations} (PDEs), discretization in time and space form the foundational algorithmic structure of scientific computing~\cite{hildebrand1987introduction,conte2017elementary,kutz2013data}.  
Discretization is required to accurately resolve all relevant spatial and temporal scales in order to produce a high-fidelity representation of the  dynamics.  
Such resolution can be prohibitively expensive, as resolving physics on fast time scales limits simulation times and the ability to model slow timescale processes, i.e. it results in well-known {\em numerical stiffness}~\cite{enright1975comparing,byrne1987stiff}.  
Time-stepping schemes are typically based on Taylor series expansions, which are local in time and have a numerical accuracy determined by the step size $\Delta t$.  
However, there is a rapidly growing effort to develop {\em deep neural networks} (DNNs) to learn time-stepping schemes unrestricted by local Taylor series constraints~\cite{parish2020time,regazzoni2019machine,qin2019data,lange2020fourier}.
We build on the flow map viewpoint of dynamical systems~\cite{Ying:2006,Brunton2010chaos,qin2019data} in order to learn {\em hierarchical time-steppers} ({HiTSs}) that explicitly exploit the multiscale flow map structure of a dynamical system over a disparate range of time-scales.
In levering features at different timescales, we can circumvent numerical stiffness and produce an accurate and efficient computational scheme that can provide exceptional efficiency in long-time simulation/forecasting and that can be integrated with classical time-stepping algorithms.

Numerical discretization has been extensively studied since the earliest days of scientific computing.  
Numerical analysis has provided rigorous estimates of error bounds for the diversity of discretization schemes developed over the past few decades~\cite{hildebrand1987introduction,conte2017elementary,kutz2013data}.  
Spatial discretization predominantly involves finite element, finite difference, or spectral methods.  
Multigrid methods have been extensively developed in physics-based simulations where coarse grained models must be progressively refined in order to achieve a required numerical precision while remaining tractable~\cite{mccormick1987multigrid,trottenberg2000multigrid}. 
The resulting discretized dynamics may be generically represented as a nonlinear dynamical system of the form
\begin{align}
        \frac{d}{dt}\boldsymbol{x}(t) = \boldsymbol{f}(\boldsymbol{x}(t),t)
\end{align}
in terms of a state ${\bf x}\in\mathbb{R}^D$ (typically $D\gg 1$).  
The dynamics are then integrated with a time-stepping algorithm.    
As with spatial discretization, there is a wide range of techniques developed for time-stepping, including explicit and implicit schemes, which have varying degrees of stability and accuracy.  
These schemes approximate the discrete-time flow map~\cite{guckenheimer2013nonlinear,wiggins2003introduction}
\vspace{-.1in}
\begin{align}\label{eq:flow}
    \boldsymbol{x}(t+\Delta t) = \boldsymbol{F}(\boldsymbol{x}(t),\Delta t) \triangleq \int_t^{t+\Delta t} \boldsymbol{f}(\boldsymbol{x}(\tau),\tau)\,d\tau,   
\end{align}
often through a Taylor-series expansion.  Runge-Kutta, for which the Euler method is a subset, is one of the standard time-stepping schemes used in practice.
Generically, it takes the form
\vspace{-.1in}
\begin{equation}
   {\bf x}_{n+1} \approx \tilde{\boldsymbol{F}}_{\Delta t}({\bf x}_n) \triangleq {\bf x}_n + \Delta t \sum_{j=1}^{k} b_j {\bf h}_j
   \label{eq:RK}
\end{equation}
\vspace{-.025in}
where
\vspace{-.025in}
\begin{equation}
  {\bf h}_j = \boldsymbol{f}\left({\bf x}_n + g\left(\sum_{k=1}^{j-1} \alpha_{j,k} {\bf h}_k \right),  t_n+c_j \Delta t \right)
\end{equation}
and ${\bf x}_n={\bf x}(t_n) = {\bf x}(n \Delta t)$.  
Note that the contributions ${\bf h}_j$ are hierarchically computed in the Runge-Kutta scheme.  
The weightings $b_j$, $c_j$ and $\alpha_{j,k}$ are derived from Taylor series expansions in order to minimize error.  
For instance, the classic fourth-order Runge-Kutta scheme, for which $k=4$ above, has a local truncation error of $\mathcal{O}(\Delta t^5)$, which leads to a global time-stepping error of $\mathcal{O}(\Delta t^4)$.
Euler stepping, for which $k=1$, has local and global time-stepping errors of $\mathcal{O}(\Delta t)$ and $\mathcal{O}(\Delta t^2)$ respectively.
Importantly, the error is explicitly related to the time-step $\Delta t$, making such time-discretization schemes {\em local} in nature.  

In contrast to schemes such as Runge-Kutta, that approximate the flow map with a local Taylor series, it is possible to directly construct an approximate flow map $\hat{\boldsymbol{F}}$ using DNN architectures.  
There are several approaches to modeling flow-map time-steppers using neural networks, which will be reviewed below.  
The approach taken in this work is to develop a hierarchy of approximate flow maps $\hat{\boldsymbol{F}}_j({\bf x},\Delta t_j)$ to facilitate the accurate and efficient simulation of multiscale systems over a range of time-scales; similar flow map composition schemes have been demonstrated to be highly effective for simulating differential equations without neural networks~\cite{Ying:2006,Brunton2010chaos,Luchtenburg2014jcp}.  
Flow maps also provide a robust framework for model discovery of multiscale physics~\cite{bramburger2020poincare,bramburger2020sparse}.
Various existing DNN architectures can be integrated into this hierarchical framework.  
Flow map approximations based on the Taylor series are typically only valid for small time steps, as the flow map becomes arbitrarily complex for large time steps in chaotic systems. 
However, DNN architectures are not limited by this small time step constraint, as they may be sufficiently descriptive to approximate exceedingly complex flow map functions~\cite{hornik1989multilayer}.  

Neural networks have been used to model dynamical systems for decades~\cite{gonzalez1998identification, milano2002neural}. 
They are computational models that are composed of multiple layers and are used to learn representations of data~\cite{lecun2015deep, goodfellow2016deep}. 
Due to their remarkable performance on many data-driven tasks~\cite{krizhevsky2012imagenet, tompson2014joint, mikolov2011strategies, senior2012deep, collobert2011natural, sutskever2014sequence}, various new architectures that favor interpretability and promote physical insight have been recently proposed, leading to many successful applications. 
In particular, it has been shown that neural networks may be used in conjunction with classical methods in numerical analysis to obtain discrete time-steppers~\cite{raissi2018multistep,parish2020time,regazzoni2019machine,qin2019data,rudy2019deep}. 
Other applications include reduced order modeling~\cite{pan2018data, wan2018data}, multiscale modeling~\cite{jacobsen2017multiscale, wehmeyer2018time, douglas1992mgnet, liu2020multiresolution}, scientific computing~\cite{raissi2017physics, bar2019learning, sitzmann2020implicit}, coordinate transformations~\cite{lusch2018deep, champion2019data}, attractor reconstructions~\cite{pathak2017using, lu2018attractor}, and forecasting~\cite{pan2018long, pathak2018hybrid, vlachas2018data, wiewel2019latent}. 
Neural network models are increasingly popular for two reasons. 
First, the universal approximation theorem guarantees that arbitrary continuous functions can be approximated by neural networks with sufficiently many hidden units~\cite{cybenko1989approximation}. 
Second, neural networks themselves can be viewed as discretizations of continuous dynamical systems~\cite{weinan2017proposal,weinan2019mean, ma2019machine, chang2017multi,chen2018neural, haber2017stable, poole2016exponential}, which makes them suitable for studying dynamics. 
Among all architectures, {\em recurrent neural networks} (RNNs) are natural candidates for temporal sequence modeling; however, training has proven to be especially difficult due to the notorious exploding/vanishing gradient problem~\cite{Hochreiter:91, bengio1994learning}. 
To alleviate this problem, new architectures have been proposed~\cite{hochreiter1997long, jaeger2002tutorial, pascanu2013difficulty}, for example, augmenting the network with explicit memory using gating mechanism, resulting in the {\em long short term memory} (LSTM) algorithm~\cite{hochreiter1997long}, or adding skipped connections to the network, leading to {\em residual networks} (ResNet)~\cite{he2016deep}. 

\begin{figure}[t!]
\vspace{-.4in}
    \centering
    \includegraphics[width=.9\textwidth]{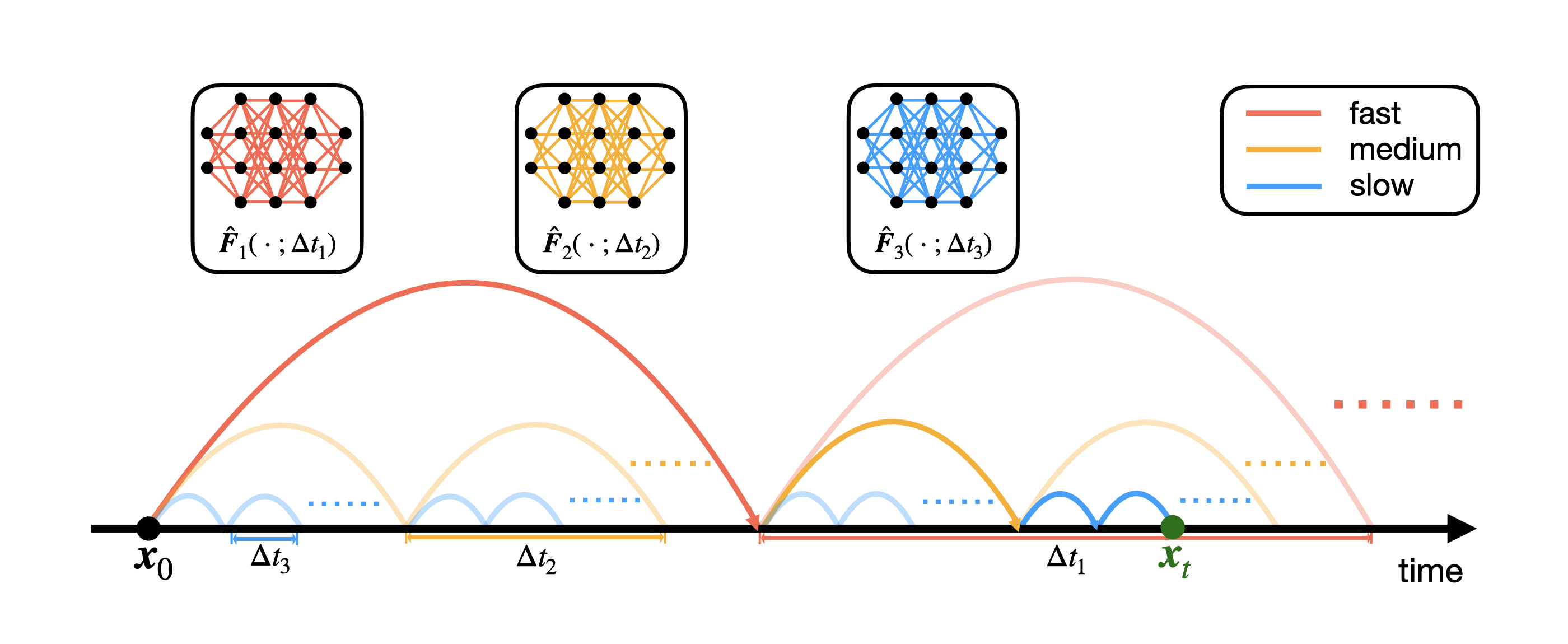}
    \vspace{-.4in}
    \caption{{\textbf{Multiscale hierarchical time-stepping scheme}}. Here, we employ neural network time-steppers over three time scales. The red model takes large steps, leaving the finer time-stepping to the yellow and blue models. The dark path shows the sequence of maps  from $\boldsymbol{x}_0$ to $\boldsymbol{x}_t$.} 
    \label{fig:diagram}
    \vspace{-.1in}
\end{figure}

In this work, we expand on~\cite{qin2019data} and employ a deep {\em residual network} (ResNet) as the basic building block for modeling the flow-map dynamics (\ref{eq:flow}).  
Unconstrained by the typical form of a Taylor-series based time-stepping scheme (\ref{eq:RK}), a multiscale modeling perspective is taken to strengthen the performance of our proposed multiscale, flow-map models. {Our multiscale HiTS algorithm} consists of
ResNet models trained to perform hierarchical time-stepping tasks. 
The contributions of this work are summarized as follows: 
\begin{itemize} \setlength\itemsep{-.05em}
    \item We propose a novel method to couple {neural network HiTSs} trained across different time scales, shown in Fig.~\ref{fig:diagram}, resulting in more accurate future state forecasts without losing computational efficiency.
    \item {Neural network HiTSs} may be coupled with classical numerical time-steppers. This hybrid time-stepping scheme can be naturally parallelized, accelerating classical numerical simulation algorithms.
    \item By coupling models across different scales, each individual model only need to be trained over a short period without being exposed to the exploding/vanishing gradient problem, enabling faster training.
    \item Despite the structural simplicity, the coupled model can still be used to capture long-term dependencies, achieving state-of-the-art performance on sequence generation.
\end{itemize}

\noindent The paper is organized as follows.
We motivate the proposed approach and present the methodology in Sec.~\ref{sec:method}. Our approach is then tested on several benchmark problems in Sec.~\ref{sec:exp}. In Sec.~\ref{sec:conclusion}, we conclude and discuss future directions. Our code is publicly available at \url{https://github.com/luckystarufo/multiscale\_HiTS}.

\section{Multiscale Time-Stepping with Deep Learning}
\label{sec:method}

\begin{figure}[t!]
    \centering
    \includegraphics[width=160mm, height=55mm]{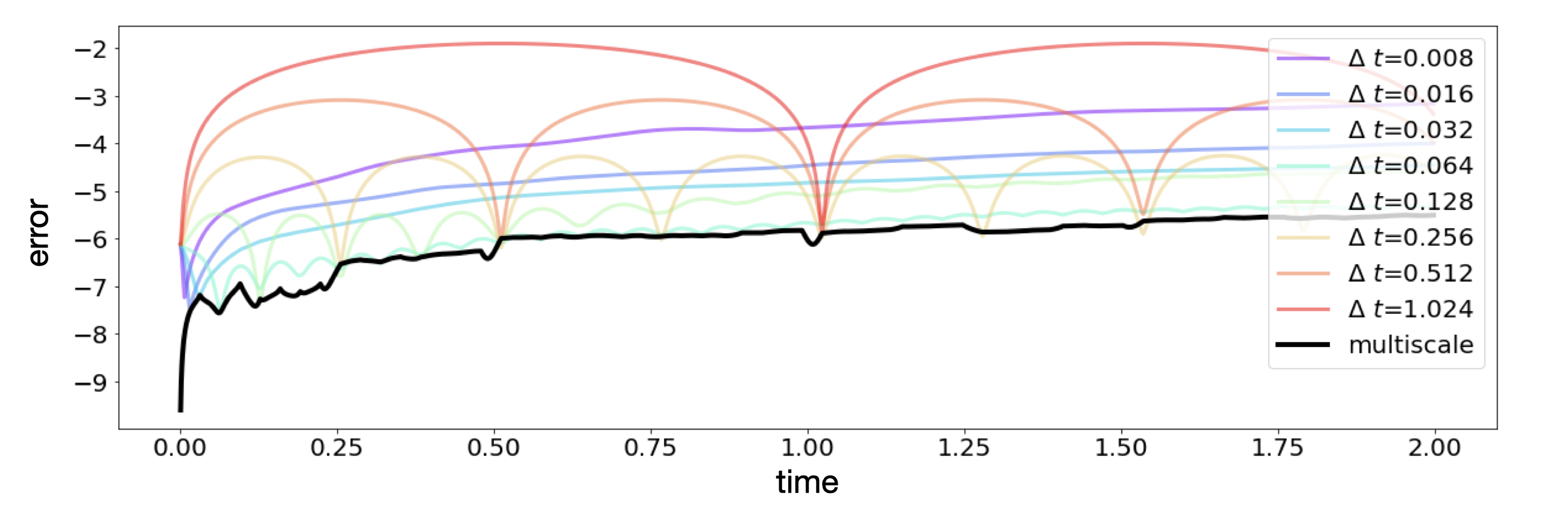}
    \vspace{-.2in}
    \caption{\textbf{Performance of {multiscale HiTS} on harmonic oscillator example}. This figure shows the time-stepping performance of different neural network time-steppers. $100$ testing trajectories are used for benchmarking each time-stepper and the mean squared errors at each step are plotted in the base-10 logarithmic scale. The black curve represents our proposed multiscale scheme whereas other colors represent time-steppers at particular scales.}
    \label{fig:linear}
\end{figure}

Here we outline our {multiscale hierarchical time-stepping} based on deep learning, illustrated in Fig.~\ref{fig:diagram}. 
Our approach constructs a hierarchy of flow maps, $\hat{\boldsymbol{F}}_j({\bf x},\Delta t_j)$, each approximated with a deep neural network.  
This enables accurate and efficient simulations with fine temporal resolution over long time scales, as compared in Fig.~\ref{fig:linear}.  
We begin with a motivating example, followed by a summary of notation and description of the training data.  Then we will introduce our {multiscale hierarchical time-stepping scheme}, including descriptions on how to vectorize operations and create hybrid time-steppers by combining with classical numerical methods.  

\subsection{Motivating example}
\label{sec:motiv_example}
To explore the effect of time step size on simulation performance, we consider the following simple linear differential equation for a harmonic oscillator
\begin{subequations}
\begin{align}
        \dot{x} &= y \\
        \dot{y} &= -x.
\end{align}
\end{subequations}
We individually train $8$ neural networks with step sizes $\Delta t$ of $0.008, 0.016, 0.032, 0.064, 0.128, 0.256, 0.512$ and $1.024$ on $500$ sampled trajectories and test them on $100$ new trajectories.  
Linear interpolation is used to estimate the state at time steps that are not directly obtained from the neural network time-stepping scheme. 
To evaluate the forecasting performance, we calculate the averaged error at each time step against the ground truth analytical solution.  
In this experiment, training and testing data are both sampled from the region $\{(x, y)| \ x^2 + y^2 \leq 1\}$, and we only train one step forward for each neural network time-stepper. 

Results are plotted in \cref{fig:linear}, where it is clear that the proposed multiscale time-stepper outperforms all fixed step models.  
Networks equipped with small time steps offer accurate short-term predictions, although error accumulates at each step and quickly dominates.  
Networks with large time steps can handle long-term predictions, although they fail to provide information between steps. 
Time-steppers with some intermediate step sizes (eg. $\Delta t=0.064$) perform well in general as they balance these two factors. 
However, by leveraging all time steppers across each scale, it is possible to create a {multiscale HiTS}, given by the black curve, that is both accurate and efficient over long time scales and with fine temporal resolution.  
Details of the methodology will be presented in the remainder of this section.

\subsection{Notation and training data}
\label{sec:notation}
For training data, we collect $n$ trajectories sampled at $p$ instances with time step $\Delta t$:
\begin{equation}
    \label{eq:dataset}
    S^{(i)} = \{(\boldsymbol{x}^{(i)}_t, \boldsymbol{x}^{(i)}_{t+\Delta t}, \boldsymbol{x}^{(i)}_{t+2\Delta t}, \cdots, \boldsymbol{x}^{(i)}_{t+p\Delta t})\}
\end{equation}
for $i = 0,1,\cdots, n-1$. 
This data is used to train neural network flow map approximations, $\hat{\boldsymbol{F}}_j({\bf x},\Delta t_j)$, for $j=0,1,\cdots, m-1$.   

We follow \cite{qin2019data} and employ the residual network as our fundamental building block of our deep learning architecture.  
Residual networks were first proposed in \cite{he2016deep} and have gained considerable prominence since. 
Specifically, our network only models the difference between time steps $\boldsymbol{x}_{n+1}$ and $\boldsymbol{x}_n$, so that it is technically the flow map minus the identity: 
\begin{equation}
    \hat{\boldsymbol{x}}_{t+\Delta t_j} = \boldsymbol{x}_t + \hat{\boldsymbol{F}}_j({\bf x},\Delta t_j)
\end{equation}
where 
\begin{equation}
    \hat{\boldsymbol{F}}_j({\bf x},\Delta t_j) = \sigma_M\left( \mathbf{A}_{M-1}\left(\cdots\sigma_1\left(\mathbf{A}_0\right)\cdots\right)\right)
\end{equation}
is a feed-forward neural network.  
The network is parameterized by the linear operators $\mathbf{A}_j$, and $\sigma_j$ are nonlinear activation functions that are chosen to be rectified linear units (ReLU). 
The extra addition creates a skipped connection from the inputs to the outputs. Here, the architecture $\hat{\boldsymbol{F}}_j$ learns the increment in states between each $\Delta t_j$ step. 
It is possible to compose the networks to take multiple steps forward in time: $\hat{\boldsymbol{F}}_j^{(k)} = \underbrace{\hat{\boldsymbol{F}}_j \circ \hat{\boldsymbol{F}}_j \circ \cdots \circ \hat{\boldsymbol{F}}_j}_{k \ \text{times}}$.
Finally, we formulate our training objective function as 
\begin{equation}
    MSE = \frac{1}{np}\sum_{i=1}^{n}\sum_{k=1}^{p} (\hat{\boldsymbol{x}}_{t+k\Delta t_j}^{(i)} - \boldsymbol{x}_{t+k\Delta t_j}^{(i)})^2
\end{equation}
which is the classical mean squared loss function.

\subsection{{Multiscale hierarchical time-stepping scheme}}
\label{sec:method_multiscale}

Multiscale modeling is ubiquitous in modern physics-based simulation models.  
Computational challenges arise in these simulations since coarse-grained macroscale models are usually not accurate enough and microscale models are too expensive to be used in practice~\cite{E:2011}. 
By coupling macroscopic and microscopic models, we hope to take advantage of the simplicity and efficiency of the macroscopic models, as well as the accuracy of the microscopic models~\cite{weinan2003multiscale}. 
Many efforts have been made towards this goal, resulting in many algorithms that exploit multiscale structure in space and time, including the multi-grid method~\cite{brandt1977multi}, the fast multipole method~\cite{greengard1997fast}, adaptive mesh refinement~\cite{berger1989local}, domain decomposition~\cite{toselli2006domain}, multi-resolution representation~\cite{daubechies1992ten}, multi-resolution dynamic mode decomposition~\cite{kutz2016multiresolution}, etc. 
Mathematical algorithms such as heterogeneous multiscale modeling (HMM)~\cite{weinan2011principles, weinan2007heterogeneous} and the equation-free approach~\cite{kevrekidis2003equation} attempt to develop general guidelines and provide principled methods for this field. 
In this work, however, we develop data-driven models using neural networks, which have different considerations than physics-based simulation models.  Regardless, the goal is still to produce accurate and efficient computational models. 

Coupling neural network time-steppers across different time scales is rather straightforward, as shown in~\cref{fig:diagram}. 
One can clearly see the time-steppers with small $\Delta t$ are responsible for the accurate time-stepping results over short periods, while the models with larger $\Delta t$ steps are used to 'reset' the predictions over longer periods, preventing error accumulations from the short-time models.  
There are additional benefits of this multiscale coupling in time. 
First, training each individual network is simpler, as it is possible to use trajectories with small $p$ so that each model may focus on its own range of interest, circumventing the problem of exploding/vanishing gradients. 
Second, the framework is flexible, so that for forecasting it is possible to vectorize the computations or utilize parallel computing technologies, enabling fast time-stepping schemes. 
Moreover, it can be easily combined with classical numerical time-steppers, resulting in hybrid schemes, boosting the performance of simulation algorithms. 

The proposed multiscale coupling procedure may resemble those used in multiscale simulations (e.g., HMM).  
However, the data-driven microscopic cannot provide accurate long-time forecasts on its own, so the coupling here is not only for efficiency but also to improve the long-time fidelity of the model.  
It should also be noted that before coupling neural network time-steppers across different time scales, cross validation is used to filter the models. 
This is practically helpful because qualities of different models may vary and we want to couple the best set of models. 
Specifically, suppose we have $m$ neural network models $\{\hat{\boldsymbol{F}}_0, \hat{\boldsymbol{F}}_1, \cdots, \hat{\boldsymbol{F}}_{m-1}\}$, ordered by their associated step sizes. 
We first determine the upper bound index $u$ so that ensembling $\{\hat{\boldsymbol{F}}_0, \hat{\boldsymbol{F}}_1, \cdots, \hat{\boldsymbol{F}}_u\}$ has the best time-stepping performance among $\{\hat{\boldsymbol{F}}_0, \hat{\boldsymbol{F}}_1, \cdots, \hat{\boldsymbol{F}}_k\}$ for all $k$. Next, we seek a lower bound index $l$ so that $\{\hat{\boldsymbol{F}}_l, \hat{\boldsymbol{F}}_{l+1}, \cdots, \hat{\boldsymbol{F}}_u\}$ performs best among $\{\hat{\boldsymbol{F}}_k, \hat{\boldsymbol{F}}_{k+1}, \cdots, \hat{\boldsymbol{F}}_u\}$ for all $k$. 

\vspace{-.075in}
\paragraph{Vectorization.}

\begin{figure}[t!]
\centering
\vspace{-.5in}
    \includegraphics[width=.7\textwidth]{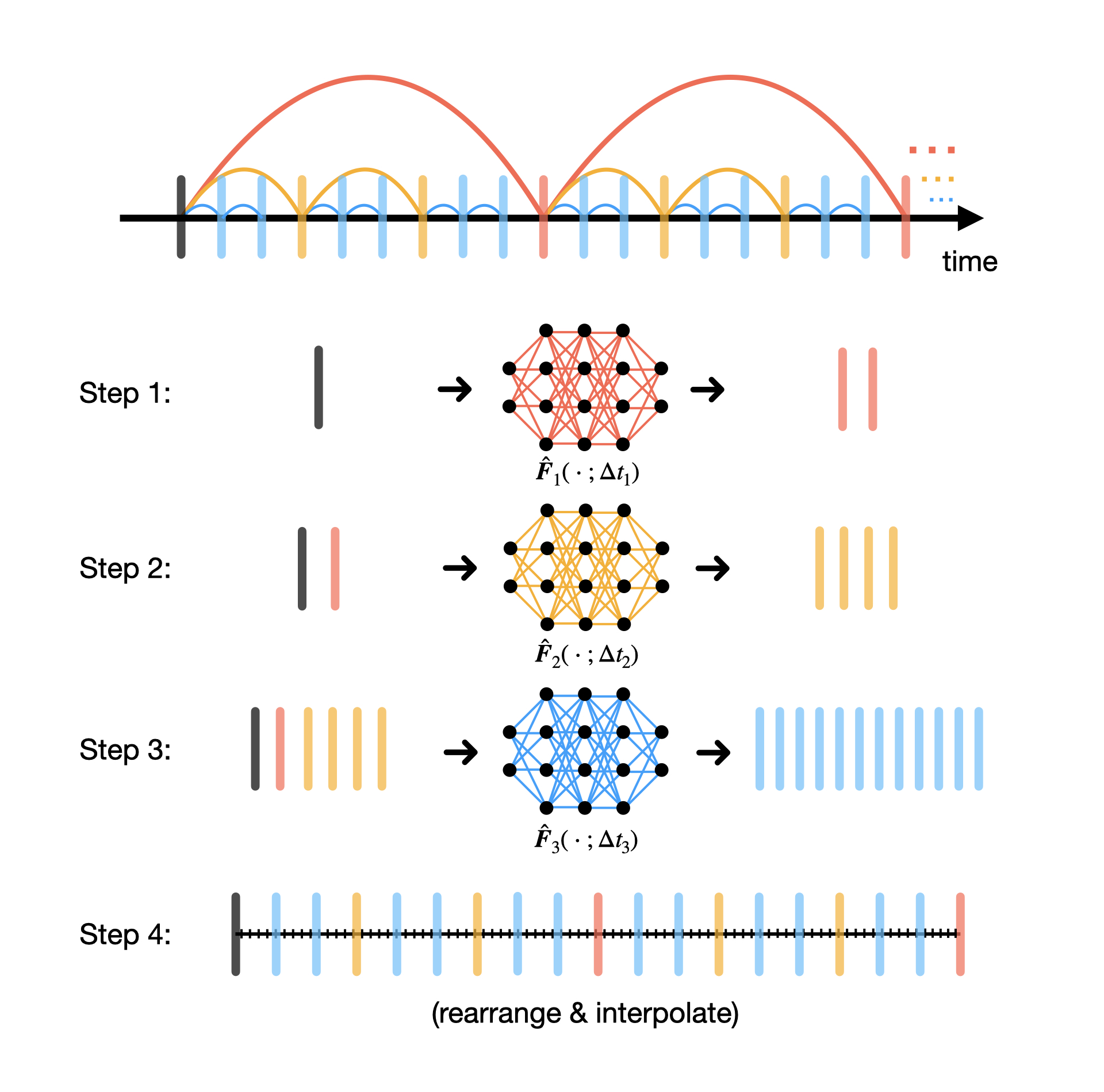}
    \vspace{-.4in}
    \caption{\textbf{Vectorized computation}. The three neural networks in this diagram are used sequentially, ordered by their associated step sizes from large to small. For each network, we stack all currently existing states and step forward (in the beginning, we only use the initial state), resulting in vectorized computations. These newly generated states are further fed to the next neural network in queue. Once we finish using all networks, the states will be rearranged in terms of chronological order and intermediate time steps will be obtained via interpolation.}
    \label{fig:vectorization}
\end{figure}

\begin{algorithm*}[t]
    \caption{{Vectorized multiscale hierarchical time-stepping}}
    \label{alg:vecorization}
    \begin{algorithmic}[1]
    \INPUT{a set of neural network time-steppers $\hat{\boldsymbol{F}}s$}
    \OUTPUT{a list of predicted states $Xs$ sorted in chronological order}
    \Function{VectorizedMultiScaleTimeStepping}{$\hat{\boldsymbol{F}}s$}
        \State $Sort(\hat{\boldsymbol{F}}s)$;   \Comment{models are ordered with decreasing step sizes}
        \State $Xs=List()$;   \Comment{create an empty list}
        \State $Append(\boldsymbol{x}_0, Xs)$;   \Comment{append initial state to the list}
        \For{i in $0, 1, \cdots, m-1$:}
            \State $K_i$ := the number of steps forward with $\hat{\boldsymbol{F}}_i$;
            \State $X_{cur} = Stack(Xs)$; \Comment{stack together all states in $Xs$}
            \State $X_{next} = \hat{\boldsymbol{F}}_i.Forward(X_{cur}, K_i)$;  \Comment{step forward $K_i$ steps with $\hat{\boldsymbol{F}}_i$ and $X_{cur}$}
            \State $Append(X_{next}, Xs)$;  \Comment{update the list with new states}
        \EndFor
        \State $Rearrange(Xs)$;
        \State $Interpolate(Xs)$;
        \Return $Xs$
    \EndFunction
    \end{algorithmic}
\end{algorithm*}

The diagram for vectorized computations is illustrated in \cref{fig:vectorization}. The basic idea is to start by using the time-steppers with the largest $\Delta t$ step and generate the future states corresponding with this stepper. We then stack the new states with the original states and feed them to the next-level neural network time-stepper. 
After we proceed through all time-steppers, we rearrange the states and use interpolation to fill in the state at all intermediate time steps. Details are given in \cref{alg:vecorization}.

\begin{algorithm*}[t]
    \caption{Hybrid time-stepping scheme}
    \label{alg:hybrid}
    \begin{algorithmic}[1]
    \INPUT{a set of neural network time-steppers $\hat{\boldsymbol{F}}s$}
    \OUTPUT{a list of predicted states $Xs$ sorted in chronological order} 
    \Function{HybridTimeStepping}{$\hat{\boldsymbol{F}}s$}
        \State $Sort(\hat{\boldsymbol{F}}s)$;  \ (models are ordered with decreasing step sizes)
        \State $Xs=List()$;   \Comment{create an empty list}
        \State // Use neural network time-steppers for large steps
        \State $Append(\boldsymbol{x}_0, Xs)$;   \Comment{append initial state to the list}
        \For{i in $0, 1, \cdots, q-1$:} \Comment{$q$ is the number of HiTSs to use}
            \State $K_i$ := the number of steps forward with $\hat{\boldsymbol{F}}_i$;
            \State $X_{cur} = Stack(Xs)$;  \Comment{stack together all states in $Xs$}
            \State $X_{next} = \hat{\boldsymbol{F}}_i.Forward(X_{cur}, \hat{\boldsymbol{F}}_i)$;  \Comment{step forward $K_i$ steps with $\hat{\boldsymbol{F}}_i$ and $X_{cur}$}
            \State $Append(X_{next}, Xs)$;  \Comment{update the list with new states}
        \EndFor
        \State // Use classic numerical time-steppers for fine steps
        \State $K$ := the number of steps forward with runge-kutta time stepper;
        \State $X_{cur} = Stack(Xs)$; \  (stack together all states in $Xs$)
        \State $X_{next} = RK4(X_{cur}, K)$;  \Comment{step forward $K$ steps with runge-kutta time-stepper and $X_{cur}$}
        \State $Append(X_{next}, Xs)$;  \Comment{update the list with new states}
        \State $Rearrange(Xs)$;
        \Return $Xs$
    \EndFunction
    \end{algorithmic}
\end{algorithm*}

\begin{figure}[t]
    \centering
    \vspace{-.4in}
    \includegraphics[width=.7\textwidth]{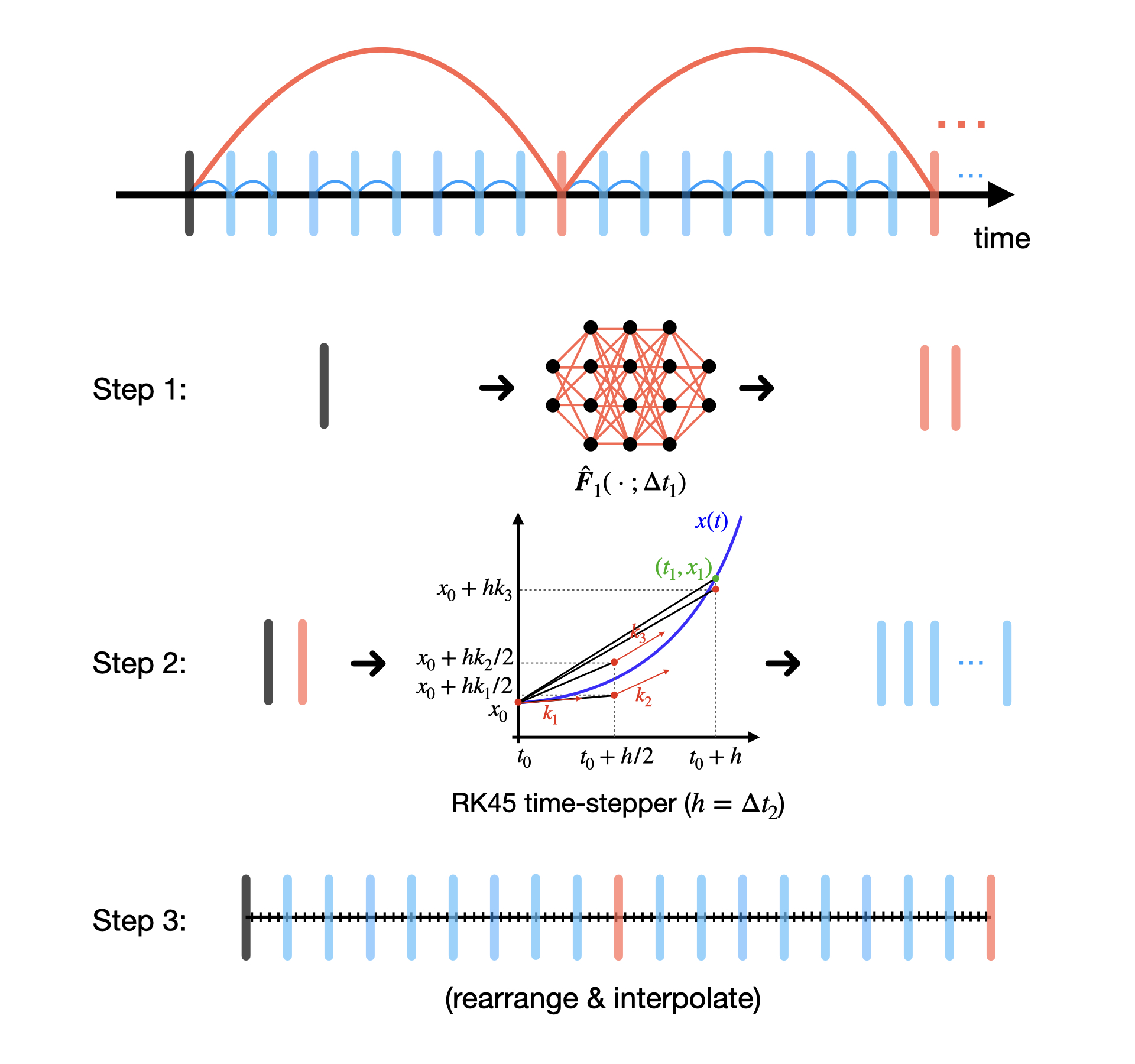}
    \vspace{-.3in}
    \caption{\textbf{Hybrid time-stepper}. A hierarchy of coarse neural network time-steppers generate states that are fed to a fourth order Runge-Kutta solver for fine-scale time-stepping.}
    \label{fig:hybrid}
\end{figure}

\vspace{-.075in}
\paragraph{Hybrid time-steppers.}
The flexibility of our {multiscale coupling approach} makes it possible to combine these time-steppers with classical numerical time-stepping algorithms. 
The algorithm is detailed in \cref{alg:hybrid} and the concept is illustrated in \cref{fig:hybrid}. 
This approach provides an innovation in the computational paradigm of numerical simulations. 
If one were to use classical numerical time-steppers (eg. Runge-Kutta method) alone, opportunities for vectorized computations or parallel computations are limited due to the serialized nature: one cannot march forward to the future without knowledge about the past.  Our hybrid scheme, on the other hand, bypasses the difficulty by utilizing large scaled neural network time-steppers, enabling vectorized computations (or parallel computations) at the bottom level where we use classical numerical time-steppers for accurate simulations. 

\section{Numerical Experiments}
\label{sec:exp}
We will now provide a thorough exploration of our proposed {multiscale hierarchical time-steppers}.  
We begin by comparing against {single time-scale neural network time-steppers} and Runge-Kutta algorithms on simple dynamical systems. We then explore this approach on more sophisticated examples in spatiotemporal physics and sequence generation.   

\subsection{Benchmark on time-stepping}
\label{sec:exp_multiscale}

\begin{figure}[ht!]
    \centering
    \includegraphics[width=\textwidth]{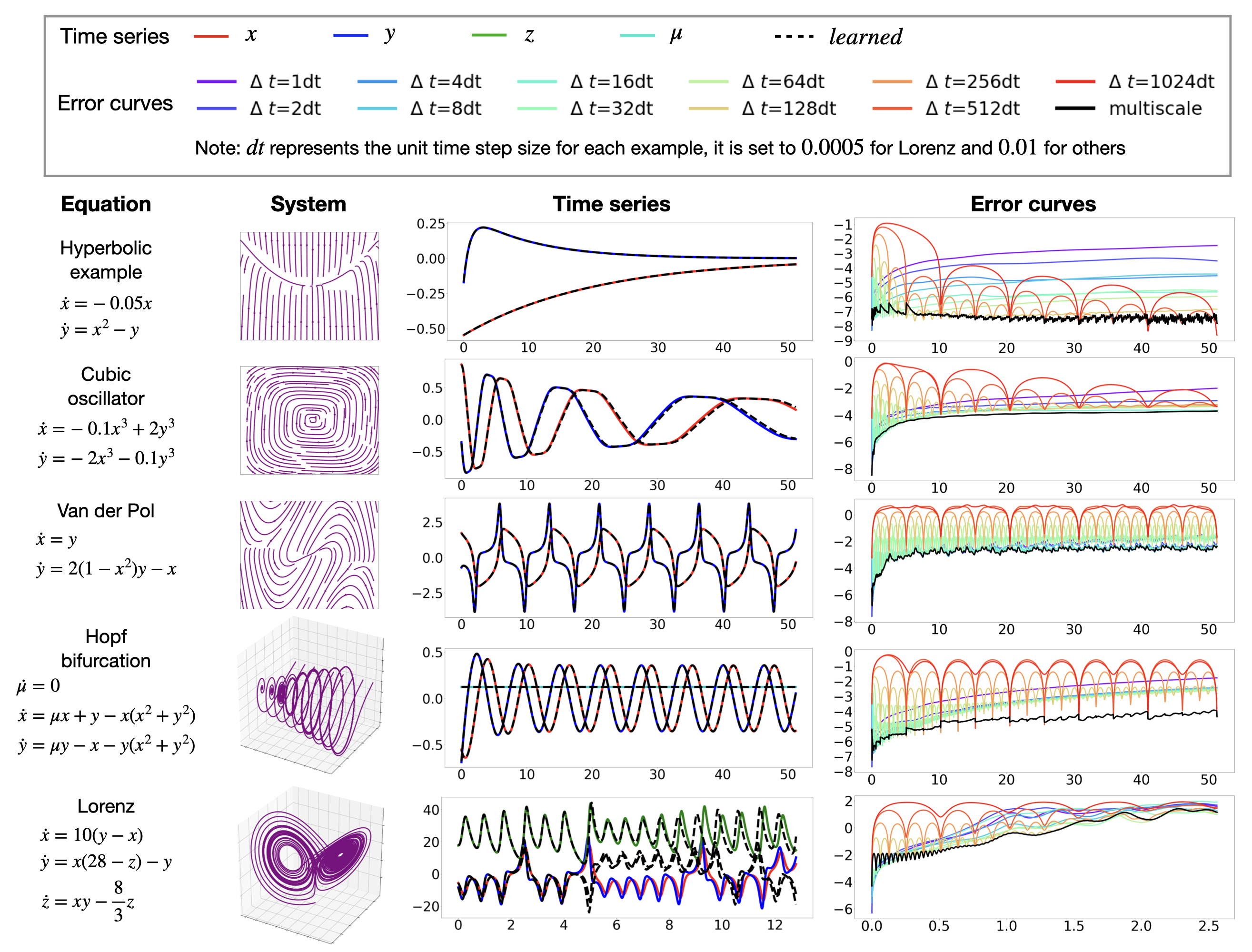}
    \caption{\textbf{Performance of {multiscale HiTSs} on nonlinear systems}. In the first two columns, systems of equations and phase portraits are visualized. In the third column, we visualize the predictions of multiscale time-steppers on a testing trajectory. In the last column, mean squared errors at each step are visualized in the base-10 logarithmic scale for different time-steppers.}
    \label{fig:nonlinear}
\end{figure}

\begin{figure}
    \centering
    \includegraphics[width=\textwidth]{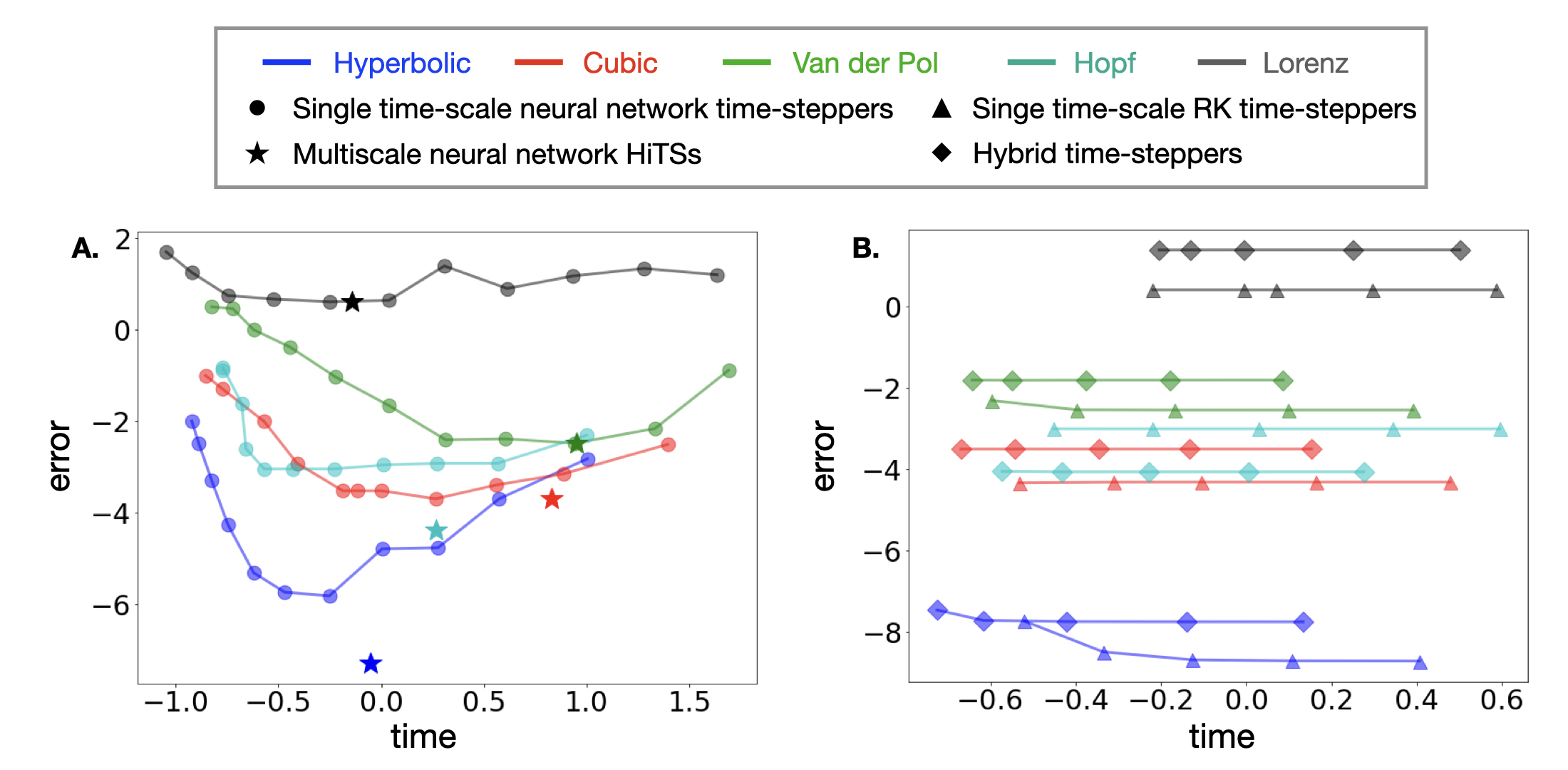}
    \caption{\textbf{Accuracy vs computational efficiency plot}. \textbf{A}. Comparison between multiscale neural network time-stepper and all single time-scale neural network time-steppers. \textbf{B}. Comparison between hybrid time-stepper and Runge-Kutta time-steppers with uniform step sizes. In both plots, horizontal and vertical axes represent time and integrated $\mathcal{L}_2$ error respectively, visualized in the base-10 logarithmic scale.}
    \label{fig:acc_vs_eff}
\end{figure}

We first benchmark the {multiscale neural network HiTS} against the {single time-scale neural network time-steppers} on five simple nonlinear dynamical systems: a nonlinear system with a hyperbolic fixed point, a damped cubic oscillator, the Van der Pol oscillator, a Hopf normal form, and the Lorenz system. 
For each example, we train $11$ {single time-scale neural network time-steppers} with separate time steps, and then combine them into a {multiscale neural network HiTS} with the methods described in \cref{sec:method_multiscale}. 
More details about these numerical experiments can be found in \cref{sec:setup_ms}.

\cref{fig:nonlinear} shows that the multiscale scheme outperforms all {single time-scale schemes} in terms of accuracy, as shown by the black curve in the last column. 
On the sampled testing trajectory, in the third column, our multiscale scheme achieves nearly perfect time-stepping for a time period of $51.20$ for the first four nonlinear systems. 
For the Lorenz system, discrepancy occurs in the forecast after about $5$ time units, when the trajectory switches lobes. 
Indeed, the error plot suggests the time-stepper becomes unreliable even after a single time unit, due to the intrinsic chaotic dynamics; more details are discussed in \cref{sec:flowMaps}. 

The trade-off between computational accuracy and efficiency are visualized in \cref{fig:acc_vs_eff}A. Here, we report the $\mathcal{L}_2$ error, averaged over all time steps and test trajectories. 
The {single time-scale scheme} curves have a ``U" shape for each example, indicating that accuracy first improves and then deteriorates as we spend more time in the computation. 
This finding is consistent with~\cite{raissi2018multistep}, which states that there is a problem-dependent sweet-spot for the hyperparameter $\Delta t$. 
Our multiscale HiTS always achieves the best accuracy, usually with a reasonable computational efficiency, due to the vectorized computations of array programming. 
For the cubic oscillator, Van der Pol oscillator, and Lorenz system, there seems to exist a single-scale neural network time-stepper with higher efficiency and competitive accuracy compared to our multiscale scheme. 
This is due to the greedy method we use for the cross validation process: we always prefer models with higher accuracy at the cost of efficiency. However, one can adjust the balance between accuracy and efficiency depending on the objective. 
The hyperparameter tuning of $\Delta t$ is implicitly conducted in the cross validation process before coupling the various scales.

Though the multiscale scheme improves the computational efficiency of neural network time-steppers, they still cannot easily match the efficiency of most classic numerical algorithms (see \cref{sec:benchmark_rk45} for more details). 
Indeed, evaluating a neural network model (forward propagation) typically requires more computational effort than applying a classic discretization scheme that only involves a few evaluations of the known vector field. 
A hybrid time-stepper, on the other hand, may break this bottleneck by combining large-scale neural network time-steppers with classic numerical time-steppers to make the computations inherently parallelizable. Therefore, we benchmark the hybrid time-steppers against the classic numerical time-steppers in~\cref{fig:acc_vs_eff}B. In particular, we use a fourth-order Runge–Kutta integrator, with $7$ different sizes of uniform time step. For simplicity, the hybrid time-steppers are constructed from the same fourth-order Runge-Kutta (RK4) time-steppers with one large-scale neural network time-stepper at the highest level. More details can be found in \cref{sec:setup_hybrid}.

\cref{fig:acc_vs_eff}B shows the trade-off between accuracy and efficiency for our hybrid time-stepper and the classic RK4 scheme. Our hybrid time-stepper offers an efficiency gain over the Runge-Kutta time-stepper with the same minimal step size. This suggests the marginal benefits of enabling vectorized computation is potentially greater than the costs of evaluating a neural network time-stepper. But the accuracy of hybrid time-steppers are usually slightly lower than the purely numerical time-steppers with rare exceptions (e.g., the Hopf normal form example), as the global error are usually dominated by the error of neural network time-steppers, which are model agnostic and purely data-driven, limiting their ability to produce high-fidelity simulations.

\subsection{Benchmark on sequence generation}
\label{sec:exp_sequence}
In addition to integrating simple low-dimensional dynamical systems, here we show that it is possible to forecast the state of more complex, high-dimensional dynamical systems.  
In the field of machine learning, this is often termed \emph{sequence generation}.  
Importantly, we benchmark our architecture against state-of-the-art networks, including long short-term memory networks (LSTMs)~\cite{hochreiter1997long}, echo state networks (ESNs)~\cite{pathak2017using}, and clockwork recurrent neural networks (CW-RNNs).   
Here, our goal is to train different architectures that can generate the target sequence as accurately as possible. 
The sequences we explore include a simulated solution of the Kuramoto–Sivashinsky (KS) equation, a music excerpt from Bach's Fugue No. 1 In C Major,  BWV 846, a simulation of fluid flow past a circular cylinder at Reynolds number 100~\cite{taira:07ibfs,taira:fastIBPM}, and a video frame of blooming flowers. 
Within each individual experiment, the various architectures have nearly the same number of parameters; more details about the data preprocessing and choice of parameters are described in \cref{sec:setup_sg}.

\begin{figure}[t!]
    \centering
    \includegraphics[width=\textwidth]{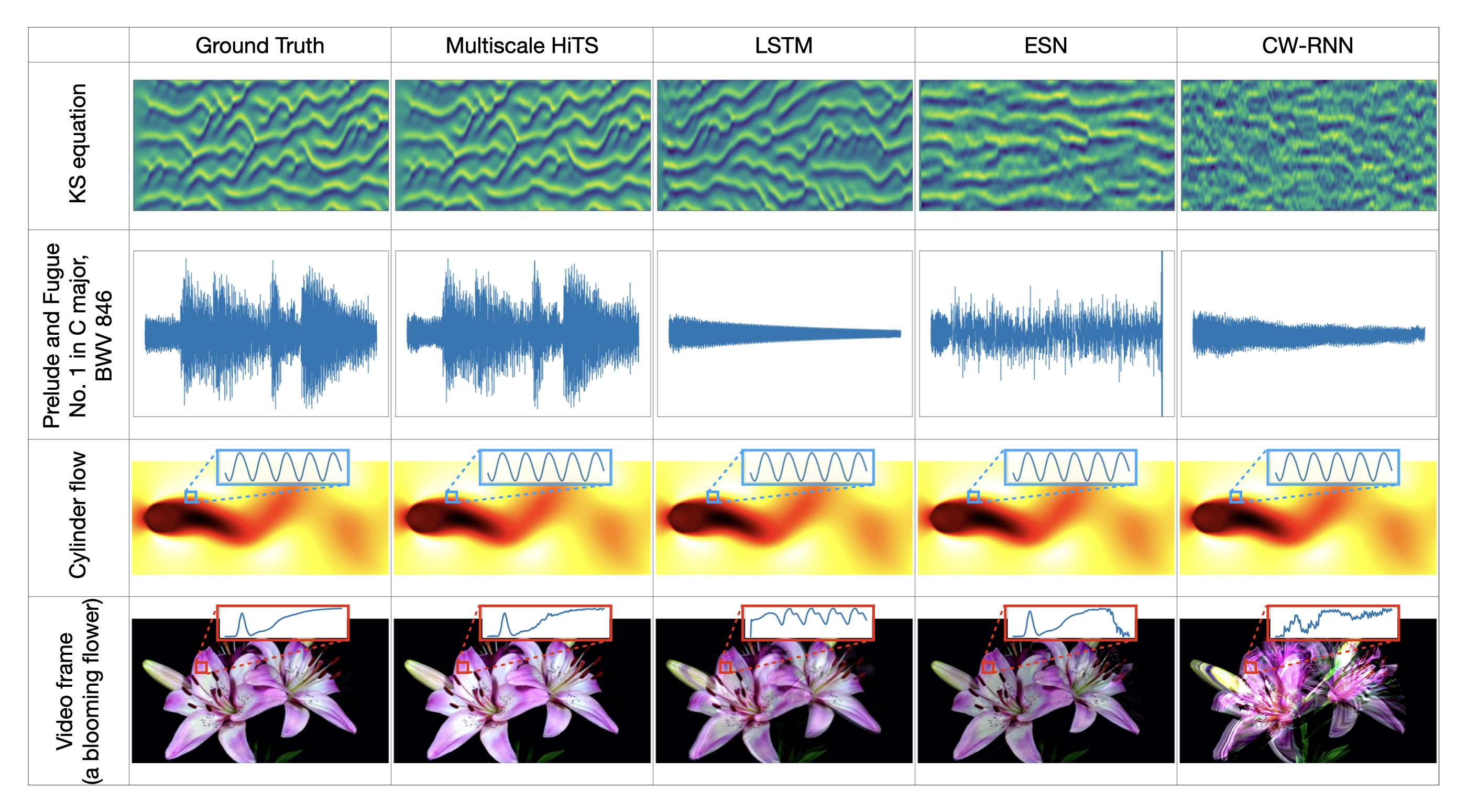}
    \vspace{-.3in}
    \caption{\textbf{Outputs of different network architectures (column) on each training sequence (row)}. We use different visualization schemes to show the results: for the KS equation and the music excerpt, we plot the time series evolution, that is, the horizontal axes represent time; For the cylinder flow and the video frame, since each state is a 2D array, we choose to visualize the last frame of our reconstruction, however, we also visualize the time evolution of some states averaged over a small patch of pixel values. For a video that shows the performance, visit:  \url{https://youtu.be/2psX5efLhCE}.}
    \label{fig:benchmarks}
\end{figure}

From \cref{fig:benchmarks}, one can visually see that the {multiscale HiTS} provides the best sequence generation results, and these results are  confirmed by the integrated $\mathcal{L}_2$ errors shown in \cref{tab:exp_benchmark}. 
We also see that the LSTM and CW-RNN can learn the first few steps accurately, whereas the ESN tends to smooth the signals. 
It should be noted that our sequence generation task are very different from the tasks considered in \cite{pathak2018model}, where they use observations of the system’s past evolution to predict future states. An ESN is usually trained by finding a set of output weights through linear regression, if the reservoir is large enough, it can in principle reproduce the observed dynamics perfectly, which would make it an inappropriate benchmark. 
An ESN with the same number of parameters as the other architectures often cannot fully describe the dynamics, resulting in coarse or smoothed approximations.  

In a nutshell, the competing architectures fail to capture the long-time behaviors, as error accumulation is inevitable for serialized computations. 
However, our proposed framework should not be viewed as a replacement for these state-of-the-art methods, as they take a different approach and philosophy for sequence generation.  RNNs tend to uncover the full dynamics with the help of memory in their internal states, whereas our scheme performs reconstruction using the time-stepping schemes learned at different time scales, though these schemes may only be accurate for a few steps.  
Instead, our multiscale framework should be used to strengthen these existing approaches, as it can utilize data-driven models across different scales, avoiding local error accumulations and potentially boosting the accuracy and efficiency.

\begin{table}[t!]
\begin{tabular}{|p{2.8cm}|p{2.8cm}|p{2.8cm}|p{2.8cm}|p{2.8cm}|}
    \hline
    \textbf{Sequences} & \textbf{HiTS} & \textbf{LSTM} & \textbf{ESN} & \textbf{CW-RNN} \\
    \hline
    \textbf{Fluid flow} &  1.23e-7 & 9.20e-8 & 2.22e-8 & 6.79e-7 \\
    \hline
    \textbf{Video frame} & 7.09e-5 & 1.44e-2 & 1.62e-3 & 8.05e-2 \\
    \hline
    \textbf{Music data} & 8.59e-7 & 4.65e-5 & 2.69e-4 & 4.70e-5 \\
    \hline
    \textbf{KS equation} & 1.04e-3 & 3.50e+0 & 3.51e+0 & 3.59e+0 \\
    \hline
\end{tabular}
\caption{The integrated $\mathcal{L}_2$ error between the generated sequence and the exact sequence.}
\label{tab:exp_benchmark}
\end{table}

\section{Conclusions \& Discussion}
\label{sec:conclusion}
In this work, we have demonstrated an effective and general data-driven time-stepper framework based on synthesizing multiple deep neural networks for hierarchical time-steppers (HiTSs) trained at multiple temporal scales.  
Our approach outperforms neural networks trained at a single scale, providing an accurate and flexible approach for integrating nonlinear dynamical systems. 
We have carefully explored this multiscale HiTS approach on several illustrative dynamical systems as well as for a number of challenging high-dimensional problems in sequence generation.  
In the sequence generation examples, our approach outperforms state-of-the-art neural network architectures, including LSTMs, ESNs, and CW-RNNs.
Our method explicitly takes advantage of dynamics on different scales by learning flow-maps for those different scales. The coupled model still maintains computational efficiency thanks to the vectorized computations of array programming. Moreover, exactly due to this coupling scheme, each individual network can focus on their intrinsic ranges of interest, bypassing the exploding/vanishing gradient problem for training recurrent neural networks. In addition, we demonstrate the joint use of our neural network time-steppers with the classical time-steppers, resulting in a new computational paradigm: numerical simulation algorithms are now parallelizable rather than serialized in nature, leading to performance boosts in computational speed. 

This work highlights fundamental differences between physics-based simulation models and data-driven models.
In the former, the error of the time-stepping constraints are determined strictly by Taylor series expansions which are local in nature and limit the time-step $\Delta t$.  The latter is a more general flow-map construction that can be trained for any time step $\Delta t$.  Thus the error is not limited by a local Taylor expansion, but rather by pairs of training data mapping the solution to a future $\Delta t$.   
However, some cautionary remarks are warranted (see \cref{sec:flowMaps}):  as with all DNN architectures, obtaining reliable large scaled time-steppers comes at the cost of significant training.  Specifically, one usually needs to acquire large enough data sets to train the appropriate deep NN architecture. In the end, we show our proposed scheme is capable of learning long-term dependencies on some more realistic data sets, achieving state-of-the-art performance.

This work also suggests a number of open questions that motivate further investigation. 
In particular, nearly every sub-field within numerical integration can be revisited from this perspective. 
For example, bringing the idea of adaptive step sizes into this framework may potentially lead to even more efficient and accurate time-stepping schemes. Although this important element may be naturally addressed, it is not yet built in to the proposed methodology in its present form. In addition, as mentioned in \cref{sec:flowMaps}, the increments of the flow maps mostly exhibit obvious multiscale features. Since deep learning is essentially an interpolation method~\cite{mallat2016understanding}, to leverage more effective training, new sampling strategies may be proposed to address this problem~\cite{dylewsky2019dynamic,bramburger2020sparse}. 
This is of crucial importance for deep learning, as neural networks are data-hungry and the curse of dimensionality makes it impossible to generate largely enough data sets for high dimensional systems. It is also important to introduce hierarchical uncertainty quantification mechanisms, as in the real world, we often have different levels of confidence across different time scales. 

\section{Acknowledgements}
SLB acknowledges funding support from the Army Research Office (W911NF-19-1-0045); SLB and JNK acknowledge funding support from the Air Force Office of Scientific Research (FA9550-19-1-0386).  
JNK acknowledges support from the Air Force Office of Scientific Research (FA9550-17-1-0329).

\appendix
\section{Methods and data}
\subsection{{Multiscale neural network HiTS}}
\label{sec:setup_ms}
\begin{table}[t!]
\vspace{-.1in}
\begin{center}
\begin{tabular}{|p{3.cm}|p{2.7cm}|p{3.3cm}|p{3.8cm}|p{2.3cm}|}
    \hline
    \textbf{Systems} & Number of samples (train / validate / test) & Sampled region $\mathcal{D}$ in state space & Network architectures & Cross-validated NNTSs \\
    \hline
    Hyperbolic & 1600/320/320 & $[-1, 1]^2$ & [2, 128, 128, 128, 2] & 3 - 10\\
    \hline
    Cubic oscillator & 3200/320/320 & $[-1, 1]^2$ & [2, 256, 256, 256, 2] & 1 - 4\\
    \hline
    Van der Pol & 3200/320/320 & $[-2, 2]\times[-4, 4]$ & [2, 512, 512, 512, 2] & 2 - 5\\
    \hline
    Hopf bifurcation & 3200/320/320 & $[-0.2, 0.6]\times[-1, 2]\times[-1, 1]$ & [3, 128, 128, 128, 3] & 2 - 11\\
    \hline
    Lorenz & 6400/640/640 & $[-0.1, 0.1]^3$ & [3, 1024, 1024, 1024, 3] & 6 - 8 \\
    \hline
\end{tabular}
\end{center}
\vspace{-.2in}
\caption{Parameters and setups for the neural network time-steppers.}
\label{tab:exp_setups}
\end{table}

The setup of our {multiscale neural network HiTS} relies on a sequence of successfully trained, single time-scale neural network time-steppers. Here, we document the working details. For all systems, we first specify a domain of interest $\mathcal{D}$ in the state space, over which we uniformly sample the initial states of training, validating, and testing trajectories. For all systems except for the Lorenz system, the time steps of the {single time-scale neural network time-steppers} are chosen to be $0.01$, $0.02$, $0.04$, $0.08$, $0.16$, $0.32$, $0.64$, $1.28$, $2.56$, $5.12$ and $10.24$. Validating and testing trajectories are different from the training trajectories and they last $51.20$ time units. 
For the Lorenz system, the process is slightly different. We simulate $3$ long trajectories to form training, validating, and testing data sets (one for each). For each of these $3$ trajectories, we cut off the initial $5$ time units to make sure the collected data are on the attractor. $11$ neural network time-steppers are trained for the Lorenz system as well; however, they have time steps of $0.0005$, $0.001$, $0.002$, $0.004$, $0.008$, $0.016$, $0.032$, $0.064$, $0.128$, $0.256$ and $0.512$ due to the inherently chaotic dynamics. The validating and testing trajectories last $2.56$ time units. 
{For convenience, we term these single time-scale neural network time-steppers NNTS 0 - NNTS 10 within each individual experiment.}
In our experiments, all trajectory data are simulated with the \textbf{odeint} function provided by the \textbf{Scipy} package and are considered as the ground truth. Upon training, the number of forward steps $p$ is set to $5$, and we use the Python API for the \textbf{PyTorch} framework and the Adam optimizer with a learning rate of $1e-3$. The training ends when the maximum epoch is reached (which is set to $100000$) or when the mean squared error on one-step prediction is lower than $1e-8$. Upon evaluation, we use the procedure presented in \cref{sec:method_multiscale}, coupling the cross-validated HiTSs to perform testing. The number of training/validating/testing samples, regions of interest, network architectures, and indices of cross validated {neural network time-steppers} are shown in \cref{tab:exp_setups}.

\subsection{Hybrid time-steppers}
\label{sec:setup_hybrid}
We benchmark our proposed hybrid time-stepper against a commonly used numerical time-stepper, a fourth-order Runge-Kutta integrator with uniform step size. For the first four nonlinear system examples, we simulate with step sizes of $0.01$, $0.02$, $0.04$, $0.08$ and $0.16$. For the Lorenz system, we shrink the sizes accordingly to $0.0005$,  $0.001$, $0.002$, $0.004$ and $0.008$. For convenience, we term them RK 0 - RK 4 for each individual experiment. For simplicity, our hybrid time-steppers are constructed to be the same Runge-Kutta time-steppers (i.e. RK 0 - RK 4) associated with one single time-scale neural-network time-stepper. The step size of this neural network time-stepper is set to be $0.512$ for Lorenz and $10.24$ for others. Similarly, we term them Hybrid 0 - Hybrid 4 for convenience.

\subsection{Details for sequence generation examples}
\label{sec:setup_sg}
\begin{table}[t!]
\begin{tabular}{|p{2.8cm}|p{2.8cm}|p{2.8cm}|p{2.8cm}|p{2.8cm}|}
    \hline
    \textbf{Sequences} & \textbf{HiTS} & \textbf{LSTM} & \textbf{ESN} & \textbf{CW-RNN} \\
    \hline
    \textbf{Fluid flow} & 15388 & 15408 & 15400 & 15622 \\
    \hline
    \textbf{Video frame} & 31216 & 30976 & 31360 & 31560 \\
    \hline
    \textbf{Music data} & 1020416 & 1018368 & 1024000 & 1042328 \\
    \hline
    \textbf{KS equation} & 4069760 & 4091232 & 4070400 & 4076232 \\
    \hline
\end{tabular}
\vspace{-.1in}
\caption{Number of parameters for different architectures of the sequence generation experiments.}
\label{tab:exp_setups2}
\end{table}

\begin{table}[t!]
\begin{center}
\begin{tabular}{|p{2.8cm}|p{13.15cm}|}
    \hline
    \textbf{Sequences} & \textbf{Network architectures in {multiscale HiTS}} \\
    \hline
    \textbf{Fluid flow} & [22, 256, 22], [22, 64, 22], [22, 16, 22], [22, 4, 22]\\
    \hline
    \textbf{Video frame} & [64, 128, 64], [64, 64, 64], [64, 32, 64], [64, 16, 64] \\
    \hline
    \textbf{Music data} & [128, 2048, 128], [128, 1024, 128], [128, 512, 128], [128, 256, 128], [128, 128, 128] \\
    \hline
    \textbf{KS equation} &  [512, 2048, 512], [512, 1024 ,512], [512, 512, 512], [512, 256, 512], [512, 128, 512]\\
    \hline
\end{tabular}
\end{center}
\vspace{-.25in}
\caption{Network architectures in {multiscale HiTS} for sequence generation experiments.}
\label{tab:exp_setups3}
\end{table}

In \cref{sec:exp_sequence}, we benchmark our multiscale HiTS against long short-term memory networks (LSTMs), echo state networks (ESNs), and clockwork recurrent neural networks (CW-RNNs) on sequence generation tasks over four data sets: the Kuramoto–Sivashinsky (KS) equation, a music snippet from Fugue No. 1 In C Major, BWV 846, an animation of fluid flow passing a cylinder, and a video frame of blooming flowers.

Information for these data sets and preprocessing steps are as follows:
\begin{itemize}
    \item For the KS equation, $4001$ snapshots on $512$ evenly distributed spatial points over the interval $(0, 16\pi)$ are simulated with a spectral method. These data are used to train different network architectures.
    \item For the music data, we sample at $1102$ Hz for $2$ seconds. Then, a time-delay embedding with $128$ delays is applied to produce a richer feature space, resulting in a training tensor of size $128 \times 2077$.
    \item For the simulated fluid flow example, data is generated using the immersed boundary projection method (IBPM)~\cite{taira:07ibfs,taira:fastIBPM} at a Reynolds number of 100; details on the numerical simulation may be found in~\cite{Brunton2019book}.  
    We apply principal component analysis (PCA) to the streamwise velocity field $u$, and use the dynamics on the first $22$  modes to train the neural networks.
    \item The video frame of blooming flowers consists of $175$ image snapshots of $540 \times 960$ pixels and $3$ channels (RGB). Similar to the fluid flow data, we preprocess it by applying PCA and work with its reduced representation in the first $64$ PC dimensions.
\end{itemize}
For each sequence, we setup different types of neural network architectures so that they have about the same number of trainable parameters; see \cref{tab:exp_setups2} for the summary of number of network parameters. The specific procedures are as follows:
\begin{itemize}
    \item We first set up our multiscale neural network time-steppers. For the fluid flow and video frame examples, we use $4$ neural network time-steppers with step sizes $1$, $4$, $16$, and $64$, respectively. For the music data, we use $5$ time-steppers and the corresponding step sizes are $1$, $5$, $25$, $125$, and $625$. For the KS example, we use $5$ time-steppers with step sizes of $1$, $6$, $36$, $216$, and $1296$. For each individual example, one step size equals to the time interval between two adjacent snapshots. For simplicity, all HiTSs only have one hidden layer, and detailed network architectures are shown in \cref{tab:exp_setups3}.
    \item Once the multiscale HiTSs are trained for each data set, we calculate the total number of parameters and use this information to create the other three types of architectures by carefully choosing the number of hidden units.
    \item For the CW-RNN, it has two other hyper-parameters: the number of internal modules and their associated clock rates. We set these parameters to match those in multiscale HiTSs for fair comparison. 
\end{itemize}

\section{Additional results}

\subsection{Computation time for neural network time-steppers}
\begin{table}[t!]
\vspace{-.2in}
\centering
\begin{tabular}{|c|c|c|c|c|c|}
    \hline
    \textbf{Systems} & Hyperbolic & Cubic oscillator & Van der Pol & Hopf bifurcation & Lorenz\\
    \hline
    NNTS $0$ & $10.06s$ & $24.74s$ & $48.95s$ & $9.98s$ & $42.98s$ \\
    \hline
    NNTS $1$ & $3.75s$ & $7.68s$ & $21.48s$ & $3.69s$ & $19.01s$ \\
    \hline
    NNTS $2$ & $1.89s$ & $3.62s$ & $8.74s$ & $1.87s$ & $8.51s$ \\
    \hline
    NNTS $3$ & $1.01s$ & $1.84s$ & $4.00s$ & $1.03s$ & $4.10s$ \\
    \hline
    NNTS $4$ & $0.56s$ & $1.00s$ & $2.06s$ & $0.59s$ & $2.03s$ \\
    \hline
    NNTS $5$ & $0.34s$ & $0.77s$ & $1.09s$ & $0.37s$ & $1.09s$ \\
    \hline
    NNTS $6$ & $0.24s$ & $0.65s$ & $0.60s$ & $0.27s$ & $0.56s$ \\
    \hline
    NNTS $7$ & $0.18s$ & $0.39s$ & $0.36s$ & $0.22s$ & $0.30s$ \\
    \hline
    NNTS $8$ & $0.15s$ & $0.27s$ & $0.24s$ & $0.21s$ & $0.18s$ \\
    \hline
    NNTS $9$ & $0.13s$ & $0.17s$ & $0.19s$ & $0.17s$ & $0.12s$ \\
    \hline
    NNTS $10$ & $0.12s$ & $0.14s$ & $0.15s$ & $0.17s$ & $0.09s$ \\
    \hline
    multiscale & $0.89s$ & $6.73s$ & $7.96s$ & $1.84s$ & $0.73s$ \\
    \hline
\end{tabular}
\caption{Computation time for multiscale and all single time-scale neural network time-steppers.}
\label{tab:HiTS_time}
\end{table}
As a supplement to \cref{fig:acc_vs_eff}A, \cref{tab:HiTS_time} shows the computational time for all single time-scale neural network time-steppers along with the timings for our multiscale time-steppers. It is not surprising to see that the computation accelerates as the step size grows. The multiscale scheme is more efficient than the finest scale neural network time-stepper in use (see \cref{tab:exp_setups} for the coupled NNTSs), which suggests the benefit of vectorized computation. 

\subsection{Noisy measurements}

\begin{figure}[t!]
    \centering
    \includegraphics[width=\textwidth]{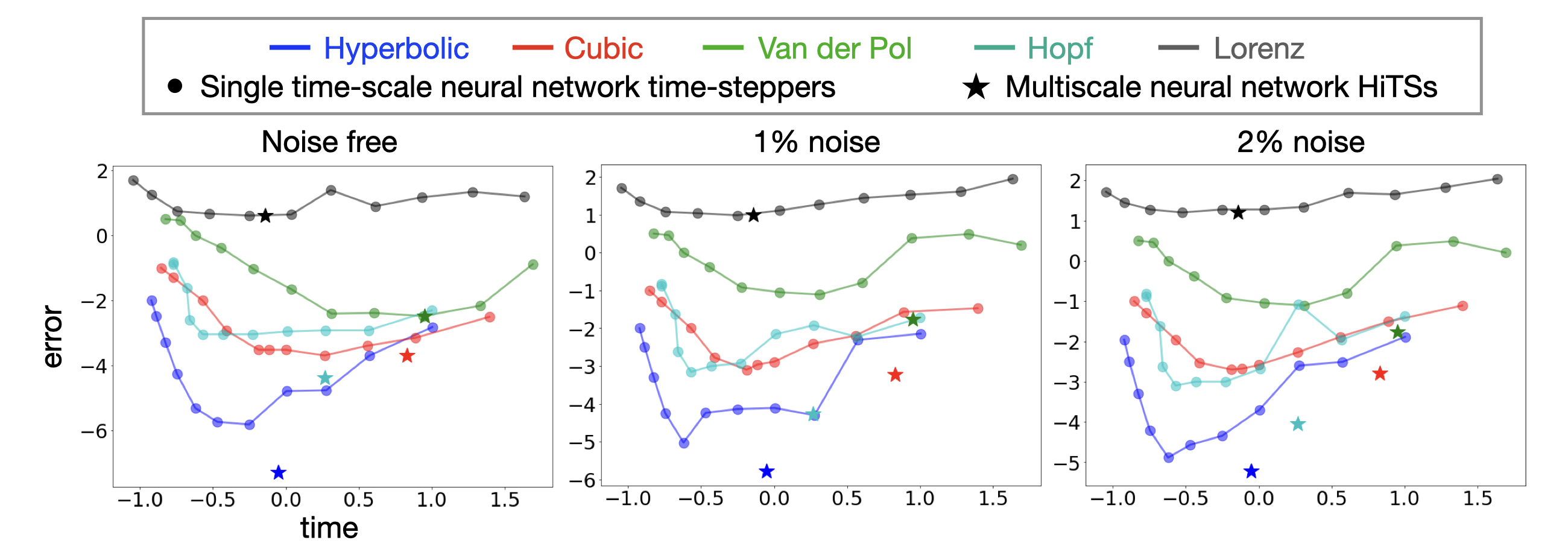}
    \vspace{-.3in}
    \caption{Accuracy and efficiency trade-offs under 0\%, 1\% and 2\% noise corruptions}
    \label{fig:noise}
\end{figure}

In \cref{tab:noise_0}, \cref{tab:noise_1}, \cref{tab:noise_2}, and \cref{fig:noise}, we study the accuracy of neural network time-steppers with different temporal gaps and the robustness of our results with respect to noise in the observations of the system states. Gaussian random noise with different variances are independently applied to each component of the dynamics. The variances are set to be $0\%$ (noise free), $1\%$, and $2\%$ of the variance of that component averaged over all trajectories across the data sets. We observe that larger noise corruption levels generally lead to inferior accuracy, but our multiscale scheme consistently works better than any {single time-scale NNTSs}. 
Similar to the results shown in \cite{raissi2018multistep}, neural networks with larger temporal gaps tend to show more robustness and the optimal value of the temporal gap is usually problem-dependent, indicating the temporal gap is a crucial parameter for data-driven modeling. By leveraging our proposed multiscale coupling strategy, it is automatically taken into account.

\begin{table}[t]
\centering
\begin{tabular}{|c|c|c|c|c|c|}
    \hline
    \textbf{Systems} & Hyperbolic & Cubic oscillator & Van der Pol & Hopf bifurcation & Lorenz\\
    \hline
    NNTS $0$ & $1.5e-3$ & $3.1e-3$ & $1.3e-1$ & $4.9e-3$ & $1.6e+1$ \\
    \hline
    NNTS $1$ & $2.0e-4$ & $7.0e-4$ & $6.9e-3$ & $1.2e-3$ & $2.2e+1$ \\
    \hline
    NNTS $2$ & $1.7e-5$ & $4.0e-4$ & $3.3e-3$ & $1.2e-3$ & $1.5e+1$ \\
    \hline
    NNTS $3$ & $1.6e-5$ & $2.0e-4$ & $4.1e-3$ & $1.1e-3$ & $8.0e+0$ \\
    \hline
    NNTS $4$ & $1.5e-6$ & $3.0e-4$ & $3.9e-3$ & $9.0e-4$ & $2.5e+1$ \\
    \hline
    NNTS $5$ & $1.8e-6$ & $3.0e-4$ & $2.2e-2$ & $9.0e-4$ & $4.4e+0$ \\
    \hline
    NNTS $6$ & $4.8e-6$ & $3.0e-4$ & $9.3e-2$ & $9.0e-4$ & $4.1e+0$ \\
    \hline
    NNTS $7$ & $5.4e-5$ & $1.2e-3$ & $4.2e-1$ & $2.5e-3$ & $4.7e+0$ \\
    \hline
    NNTS $8$ & $5.0e-4$ & $9.8e-3$ & $1.0e+0$ & $2.4e-2$ & $5.6e+0$ \\
    \hline
    NNTS $9$ & $3.2e-3$ & $5.0e-2$ & $2.9e+0$ & $1.5e-1$ & $1.8e+1$ \\
    \hline
    NNTS $10$ & $1.0e-2$ & $1.0e-1$ & $3.2e+0$ & $1.3e-1$ & $5.1e+1$ \\
    \hline
    multiscale & $5.1e-8$ & $2.0e-4$ & $3.2e-3$ & $4.1e-5$ & $4.1e+0$ \\
    \hline
\end{tabular}
\vspace{-.1in}
\caption{Integrated $\mathcal{L}_2$ between the predicted and the exact measurement with noise free data.}
\label{tab:noise_0}
\end{table}

\begin{table}[t!]
\centering
\begin{tabular}{|c|c|c|c|c|c|}
    \hline
    \textbf{Systems} & Hyperbolic & Cubic oscillator & Van der Pol & Hopf bifurcation & Lorenz\\
    \hline
    NNTS $0$ & $7.2e-3$ & $3.4e-2$ & $1.6e+0$ & $1.9e-2$ & $9.0e+1$ \\
    \hline
    NNTS $1$ & $4.9e-3$ & $2.7e-2$ & $3.1e+0$ & $5.9e-3$ & $4.1e+1$ \\
    \hline
    NNTS $2$ & $5.0e-4$ & $6.4e-3$ & $2.4e+0$ & $1.2e-2$ & $3.4e+1$ \\
    \hline
    NNTS $3$ & $7.8e-5$ & $3.9e-3$ & $1.6e-1$ & $7.0e-3$ & $2.8e+1$ \\
    \hline
    NNTS $4$ & $7.3e-5$ & $1.3e-3$ & $7.8e-2$ & $1.2e-3$ & $1.9e+1$ \\
    \hline
    NNTS $5$ & $5.8e-5$ & $1.1e-3$ & $9.0e-2$ & $1.0e-3$ & $1.3e+1$ \\
    \hline
    NNTS $6$ & $9.5e-6$ & $8.0e-4$ & $1.2e-1$ & $7.0e-4$ & $9.7e+0$ \\
    \hline
    NNTS $7$ & $5.7e-5$ & $1.7e-3$ & $4.2e-1$ & $2.4e-3$ & $1.1e+1$ \\
    \hline
    NNTS $8$ & $5.0e-4$ & $1.0e-2$ & $1.0e+0$ & $2.4e-2$ & $1.2e+1$ \\
    \hline
    NNTS $9$ & $3.2e-3$ & $5.0e-2$ & $2.9e+0$ & $1.5e-1$ & $2.3e+1$ \\
    \hline
    NNTS $10$ & $1.0e-2$ & $1.0e-1$ & $3.2e+0$ & $1.3e-1$ & $5.2e+1$ \\
    \hline
    multiscale & $1.7e-6$ & $6.0e-4$ & $1.7e-2$ & $5.4e-5$ & $9.8e+0$ \\
    \hline
\end{tabular}
\vspace{-.1in}
\caption{Integrated $\mathcal{L}_2$ between the predicted and the exact measurement with $1\%$ Gaussian noise.}
\label{tab:noise_1}
\end{table}

\begin{table}
\centering
\begin{tabular}{|c|c|c|c|c|c|}
    \hline
    \textbf{Systems} & Hyperbolic & Cubic oscillator & Van der Pol & Hopf bifurcation & Lorenz\\
    \hline
    NNTS $0$ & $1.3e-2$ & $7.8e-2$ & $1.6e+0$ & $4.2e-2$ & $1.1e+2$ \\
    \hline
    NNTS $1$ & $3.1e-3$ & $3.1e-2$ & $3.1e+0$ & $1.1e-2$ & $6.8e+1$ \\
    \hline
    NNTS $2$ & $2.5e-3$ & $1.3e-2$ & $2.4e+0$ & $8.5e-2$ & $4.5e+1$ \\
    \hline
    NNTS $3$ & $2.0e-4$ & $5.3e-3$ & $1.6e-1$ & $2.1e-3$ & $4.9e+1$ \\
    \hline
    NNTS $4$ & $4.5e-5$ & $2.6e-3$ & $7.8e-2$ & $1.0e-3$ & $2.2e+1$ \\
    \hline
    NNTS $5$ & $2.7e-5$ & $2.1e-3$ & $9.0e-2$ & $1.0e-3$ & $1.9e+1$ \\
    \hline
    NNTS $6$ & $1.3e-5$ & $2.0e-3$ & $1.2e-1$ & $8.0e-4$ & $1.9e+1$ \\
    \hline
    NNTS $7$ & $6.1e-5$ & $3.0e-3$ & $4.2e-1$ & $2.4e-3$ & $1.6e+1$ \\
    \hline
    NNTS $8$ & $5.0e-4$ & $1.1e-2$ & $1.0e+0$ & $2.4e-2$ & $1.9e+1$ \\
    \hline
    NNTS $9$ & $3.2e-3$ & $5.2e-2$ & $2.9e+0$ & $1.5e-1$ & $2.8e+1$ \\
    \hline
    NNTS $10$ & $1.1e-2$ & $1.0e-1$ & $3.2e+0$ & $1.3e-1$ & $5.2e+1$ \\
    \hline
    multiscale & $6.1e-6$ & $1.6e-3$ & $1.7e-2$ & $9.0e-5$ & $1.6e+1$ \\
    \hline
\end{tabular}
\vspace{-.1in}
\caption{Integrated $\mathcal{L}_2$ between the predicted and the exact measurement with $2\%$ Gaussian noise.}
\label{tab:noise_2}
\end{table}

\begin{table}[t!]
\centering
\begin{tabular}{|c|c|c|c|c|c|}
    \hline
    \textbf{Systems} & Hyperbolic & Cubic oscillator & Van der Pol & Hopf bifurcation & Lorenz\\
    \hline
    RK 0 & $2.55s$ & $3.18s$ & $2.82s$ & $3.93s$ & $3.87s$ \\
    \hline
    RK 1 & $1.28s$ & $1.60s$ & $1.30s$ & $2.21s$ & $1.97s$ \\
    \hline
    RK 2 & $0.74s$ & $1.01s$ & $0.78s$ & $1.07s$ & $1.17s$ \\
    \hline
    RK 3 & $0.46s$ & $0.64s$ & $0.42s$ & $0.60s$ & $0.98s$ \\
    \hline
    RK 4 & $0.30s$ & $0.43s$ & $0.30s$ & $0.35s$ & $0.60s$ \\
    \hline
    Hybrid 0 & $1.36s$ & $2.25s$ & $1.37s$ & $1.89s$ & $3.17s$ \\
    \hline
    Hybrid 1 & $0.72s$ & $0.96s$ & $0.69s$ & $1.01s$ & $1.78s$ \\
    \hline
    Hybrid 2 & $0.37s$ & $0.55s$ & $0.42s$ & $0.59s$ & $0.99s$ \\
    \hline
    Hybrid 3 & $0.24s$ & $0.30s$ & $0.29s$ & $0.37s$ & $0.74s$ \\
    \hline
    Hybrid 4 & $0.19s$ & $0.23s$ & $0.22s$ & $0.26s$ & $0.62s$ \\
    \hline
    multiscale & $0.89s$ & $6.73s$ & $7.96s$ & $1.84s$ & $0.73s$ \\
    \hline
\end{tabular}
\vspace{-.1in}
\caption{Computation time of various schemes on testing trajectories.}
\label{tab:multi_hybrid}\vspace{-.1in}
\end{table}

\subsection{More benchmark results against Runge-Kutta time-steppers}
\label{sec:benchmark_rk45}
\begin{figure}[t!]
\vspace{-.35in}
    \centering
    \includegraphics[width=\textwidth]{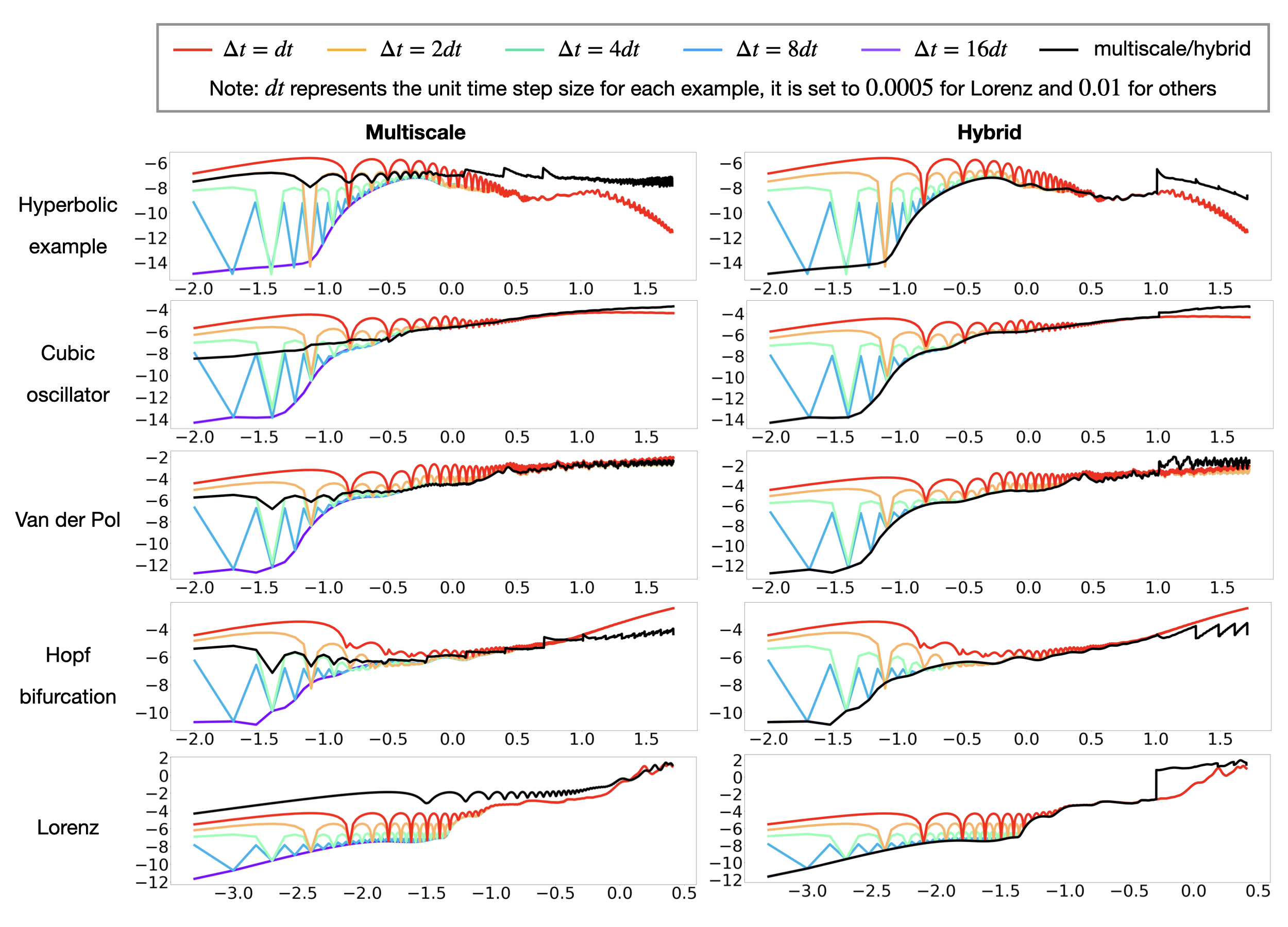}
\vspace{-.4in}    \caption{\small\textbf{Stepwise error comparisons with Runge-Kutta integration}. For each plot, the horizontal axis represents simulation time and vertical axis represents $\mathcal{L}_2$ error. Both axes are visualized in base-10 logarithmic scale. The left column shows the comparisons between multiscale time-steppers and Runge-Kutta time-steppers. The right column shows the comparisons between hybrid time-steppers and Runge-Kutta time-steppers. For both columns, mean squared errors at each time step are plotted.}
    \label{fig:rk45}
\end{figure}

\cref{fig:rk45} shows the mean squared errors of different schemes versus simulation time. Again, they are plotted in a base-10 logarithmic scale. Our multiscale schemes provide competitive results compared to the Runge-Kutta schemes in terms of accuracy. This is especially notable for the purely data-driven, neural network-based multiscale time-stepping scheme: the true model of the underlying dynamics is unavailable, which is in sharp contrast to the classical simulation setting. In other words, neural network time-steppers can identify the dynamics and generate accurate predictions at the same time. Hybrid time-steppers, on the other hand, also provide competitive accuracy. And the accuracy level is dominated by either the numerical time-stepper or the neural network time-stepper, whichever is worse. Therefore, the hybrid time-steppers usually bear a slightly lower accuracy than the purely numerical time-steppers.

\cref{tab:multi_hybrid} shows the computational times of the various algorithms which is a supplement for \cref{fig:acc_vs_eff}  For Runge-Kutta time-steppers, computations speed up as the time step increases. Both our {multiscale neural network hierarchical time-stepping scheme} and hybrid time-stepping scheme provide efficient computations. In particular, the hybrid time-steppers outperform the corresponding RK integrators, as the vectorized computations accelerate these simulations. The hybrid time-steppers generally beat the multiscale neural network time-steppers in terms of the speed. This speed-up may stem from the computation on each individual step: evaluating a large neural network model (forward propagation) requires more computational effort than applying the Runge-Kutta scheme, which only involves four evaluations of the vector field.

\subsection{Remarks on learning flow maps}
\label{sec:flowMaps}
\begin{figure}[ht!]
    \centering
    \includegraphics[width=145mm, height=200mm]{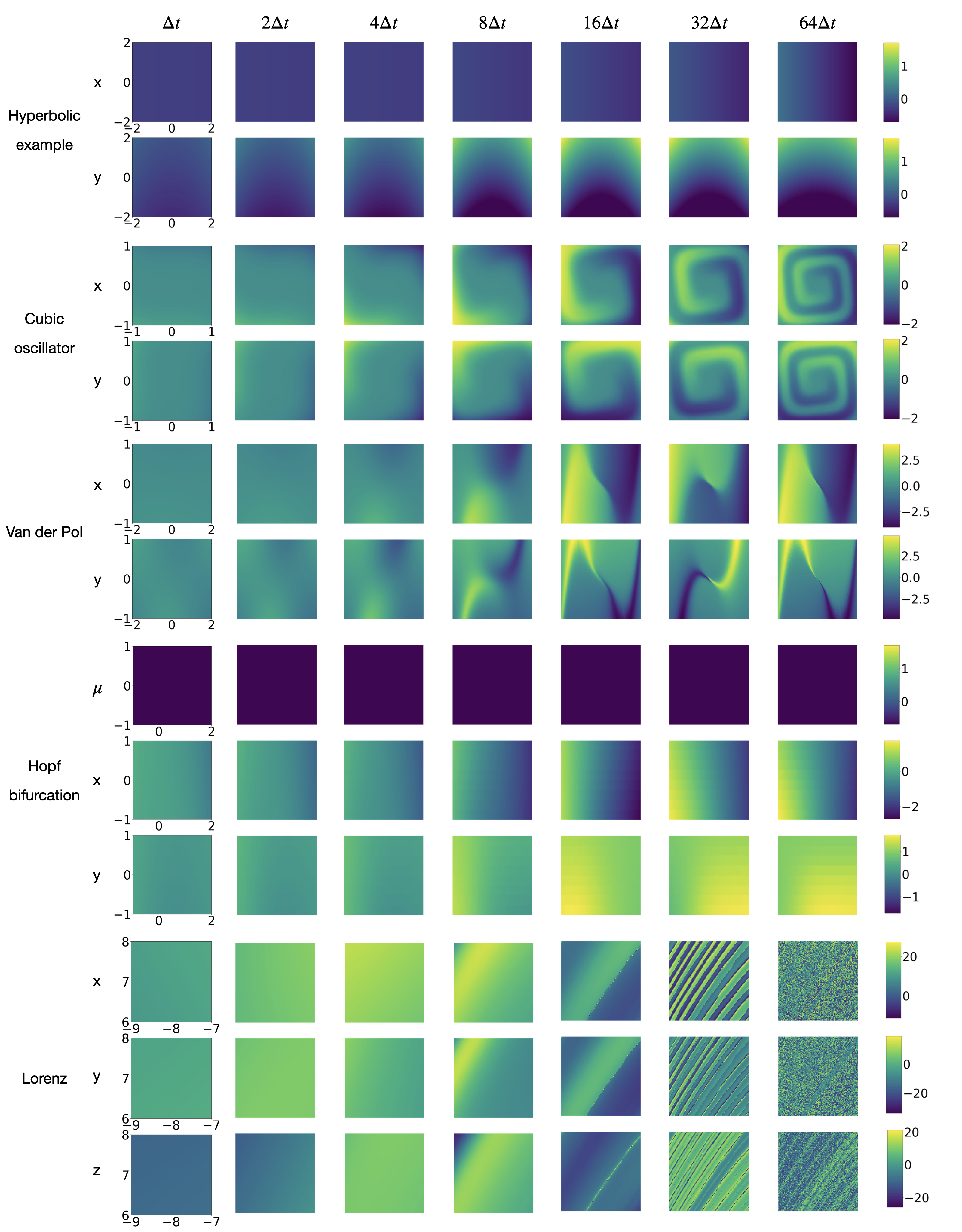}
    \vspace{-.25in}
    \caption{\textbf{Increments of flow maps}. We visualize $\boldsymbol{x}_{j\Delta t} - \boldsymbol{x}_{0}$, for $j = 1, 2, \cdots 64$ where $\boldsymbol{x}$ is the state of each example. For the Lorenz system, we set $\Delta t=0.008$ and the region we visualize is $[-9, -7]\times[6, 8] \times \{27\}$ whereas for other examples, $\Delta t=0.16$ and the regions are in \cref{tab:exp_setups}.}
    \label{fig:flowMaps}
\end{figure}

Though there is evidence that our schemes can boost the performance of neural network time-steppers, as well as classical numerical simulation algorithms, each individual result heavily depends on the system under study. 
Theoretically, there is nothing preventing us from fitting the flow maps with arbitrarily long time periods; however, this is not always possible. 
For example, in the Lorenz system, one can clearly see the performance of NNTS $10$ is unsatisfactory: the one-step prediction error is on the order of $\mathcal{O}(1)$. The reason could either be that the current architecture is not large enough to encode the map, or the training data set is not large enough to fully capture the complexity of the corresponding flow map. 
Chaotic dynamics will generally present this challenge, as the flow map complexity grows exponentially with time.  
As presented in \cref{sec:notation}, we are essentially fitting the $\Delta t-$lag flow map minus the identity with the neural network
\begin{equation}
    \hat{\boldsymbol{F}}(\boldsymbol{x}_t, \Delta t) \approx \boldsymbol{x}_{t + \Delta t} - \boldsymbol{x}_t.
\end{equation}
To see what these increments look like, for the first four nonlinear systems, we have visualized $\boldsymbol{x}_{j\Delta t} - \boldsymbol{x}_{0}$, for $j = 1, 2, \cdots 64$ with $\boldsymbol{x}_{0}'s$ uniformly sampled from their associated region on interest $\mathcal{D}$, shown in \cref{tab:exp_setups}. As for the Lorenz system, the process is similar but the initial states are sampled from the region $[-9, -7]\times[6, 8] \times \{27\}$ because this region is close to the strange attractor, and $\Delta t$ is set to $0.008$ for Lorenz and $0.16$ for others. Results are shown in \cref{fig:flowMaps}.

In the examples of hyperbolic fixed point and Hopf bifurcation, the plots do not visually grow more complex as time proceeds. For the Van der Pol oscillator, complexity grows in the beginning stage but later it tends to maintain that level of complexity. For the Cubic oscillator example, the complexity keeps growing with time; however, there are clear structures, as more spirals are generated. 
The most interesting case is the Lorenz system: not only does the complexity grow, but earlier stripe patterns (period doubling) also disappear as time proceeds, leading to tremendous difficulties in training. To fully capture this growing complexity, one may need an exponentially growing data set and network architecture. In addition, one seemingly common feature from all plots is that the multiscale effects become more pronounced as time proceeds, leading to the emergence of patterns. Whether we can utilize this feature to perform more effective training may be a promising direction that motivates future work.

\clearpage 
\begin{spacing}{.8}
 \small{
 \setlength{\bibsep}{2.9pt}
\bibliographystyle{IEEEtran}
\bibliography{ref}
 }
 \end{spacing}
\end{document}